\documentclass[conference]{IEEEtran}
\IEEEoverridecommandlockouts

\usepackage{amsmath,amssymb,amsfonts}
\usepackage{bm} 

\usepackage[ruled,vlined,linesnumbered]{algorithm2e}
\DontPrintSemicolon
\SetAlgoLined
\SetAlFnt{\footnotesize} 
\SetKwInOut{Input}{Input}
\SetKwInOut{Output}{Output}
\SetKw{Continue}{continue}

\usepackage[final]{graphicx}
\usepackage{textcomp}
\usepackage{xcolor}
\usepackage{svg}
\usepackage{url} 
\setkeys{Gin}{draft=false}

\usepackage{booktabs} 
\usepackage{tabularx} 
\usepackage{array}    
\usepackage{ragged2e} 
\usepackage{threeparttable} 
\usepackage{capt-of} 
\usepackage{dblfloatfix} 
\usepackage[section]{placeins} 

\usepackage{tikz}
\newcommand{\pill}[2]{%
  \tikz[baseline=(n.base)]\node (n) [rounded corners=3pt, fill=#1, inner xsep=5pt, inner ysep=2.2pt]{\bfseries\footnotesize #2};%
}
\newcommand{\yes}{\pill{green!18}{yes}}

\newcommand{\nc}{\pill{orange!20}{n/c}}
\newcommand{\NA}{\pill{gray!20}{n/a}}

\usepackage{biblatex}
\def\BibTeX{{\rm B\kern-.05em{\sc i\kern-.025em b}\kern-.08em
    T\kern-.1667em\lower.7ex\hbox{E}\kern-.125emX}}
\begin{document}

\title{Discovering Entity-Conditioned Lag Heterogeneity\\
{\footnotesize \textsuperscript{*}A Lag-Gated Neural Audit Framework for Panel Time Series}
}

\author{\IEEEauthorblockN{Andi Xu}
\IEEEauthorblockA{\textit{School of Engineering} \\
\textit{Jönköping University}\\
Jönköping, Sweden \\
oktinner@outlook.com}
}

\maketitle

\begin{abstract}
Country-level temporal panels are widely used in empirical analysis. Researchers often need to audit how different entities respond to historical signals over different time horizons. Current approaches typically do not provide directly auditable entity-specific lag summaries. We formulate entity-conditioned heterogeneous lag discovery as a temporal panel mining task and propose AC-GATE, an Adaptive-Conditioning Encoder with a Scale-Invariant Lag Gate. It instantiates conditional Moderated Distributed Lag by using observable entity-level proxies to condition lag-weight distributions over historical observations, thereby making effective lags structural outputs of the model rather than post-hoc explanations. The evaluation is based on a layered audit protocol that separates predictive calibration from lag discovery. A synthetic panel with known ground-truth lags is used for mechanism recovery testing, and two real-world country-level panels are used for external audit and stress testing. The results show that AC-GATE can recover heterogeneous lag structure in synthetic data, and generates non-degenerate, externally structured effective lags in real data.
\end{abstract}


\section{Introduction}
Analysts of panel time series often need more than accurate forecasts. At different entity levels (e.g., country, region, or organization), a practical question is: Do different entities respond to historical signals over systematically different time scales? Classic research and recent surveys point out that the merging of dynamic relationships across entities can be misleading and reduce explainability when response parameters differ across units \cite{pesaran1995pmg,thayasivam2025panelsurvey}. These challenges are significant when the research focuses on entity-specific effective lags in response to historical information. In productivity analysis, inputs related to technology or ability may only impact outcomes after an adjustment period, organizational learning, or complementary investments \cite{schweikl2020lessons}. In research on energy and emissions, the impacts of renewable energy, energy efficiency or institutional conditions may differ by country, development stage and emissions regime \cite{akram2020energy,mirziyoyeva2022re}. These observations indicate that single global lag structures may be too restrictive. Neural forecasting models provide flexible tools for prediction, but models such as Temporal Fusion Transformer (TFT) \cite{lim2021tft} and Gated Attention Network (GA-Net) \cite{xue2020gated} are not designed to make comparable entity-level lag distributions the primary output.

We propose AC-GATE, an Adaptive-Conditioning (AC) Encoder with a Scale-Invariant Lag Gate, which instantiates \emph{Conditional Moderated Distributed Lag} (CMDL) to model entity-conditioned heterogeneous lags in panel time series. This structure uses observable entity-level proxy information to condition (and hence moderate) the past observations, thereby constructing weight distributions over lags based on entity characteristics. The lag gate then yields an entity-level effective lag as a structural model output. To evaluate AC-GATE, we use a layered audit protocol that separates predictive calibration from lag discovery. We further include structural ablations and a proxy-shuffle negative control to distinguish architectural necessity from proxy--entity correspondence.

\begin{figure}[!tbhp]
    \centering
    \includegraphics[width=\columnwidth]{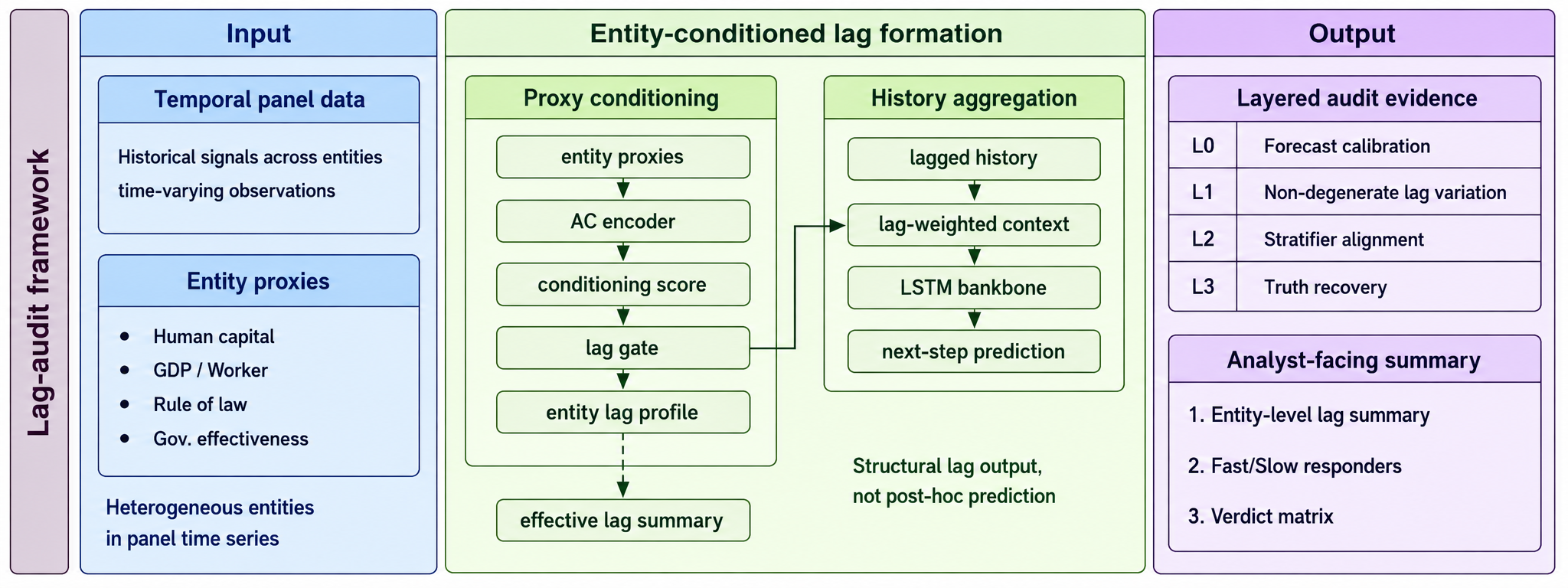}
    \caption{Use-case overview of entity-conditioned lag auditing. }
    \label{fig:use_case}
\end{figure}

This paper makes three contributions. First, we formulate entity-conditioned heterogeneous lag discovery as a testable panel time-series mining task. Second, we introduce AC-GATE, which generates entity-level effective lags via an explicit lag gating structure. Third, we propose a layered audit protocol that evaluates forecast calibration, lag non-degeneracy, external stratifier alignment and synthetic mechanism recovery. Results show that AC-GATE recovers true heterogeneous lag structure in synthetic panels and detects non-degenerate, externally structured effective-lag heterogeneity in two real-world panels. Fig.~\ref{fig:use_case} illustrates the applied lag-audit use case studied in this paper.

\section{Related Work}\label{related_work}
Panel time series data contain temporal dependence and cross-sectional heterogeneity, so they cannot be considered independent and identically distributed (i.i.d.) samples or ordinary univariate time series. The econometric literature shows that heterogeneity can display as discrete regimes or differences in distribution location or variable selection structure \cite{hansen1999thr,canay2011sim,babii2021mlpanel,Arkhangelsky2024survey}. A recent review \cite{thayasivam2025panelsurvey} also emphasizes that the core challenges of panel time-series forecasting include cross-sectional heterogeneity, temporal dependence, non-stationary and explainability. These works indicate that the lag relationships in panel data should not be considered as homogeneous across entities, which should allow for entity-conditioned dynamic response differences. Therefore, even if existing panel forecasting frameworks can deal with heterogeneity, they usually do not produce verifiable entity-specific lag structures, which is the gap that this paper aims to fill.

Deep time-series models can capture nonlinear dynamics but often lack explainability\cite{wu2023timesnet,liu2024itransformer}. The Temporal Fusion Transformer (TFT) combines static covariates, recurrent processing, attention and gating components for multi-horizon forecasting, and provides interpretability outputs \cite{lim2021tft}. On the other hand, the Gated Attention Network (GA-Net) shows that models can use input-dependent gates to select a subset of key elements, which can reduce redundant attention and improve interpretability \cite{xue2020gated}. Existing neural panel models have tried to put interpretable structure into the network, such as persistent change filters or nonlinear panel components, which can reveal auditable structure beyond forecasting \cite{yang2020panelnn,chronopoulos2023dnnpanel}. However, the TFT focuses on improving predictive performance and global explanation; GA-Net is not designed to produce a stable, cross-entity comparable lag-weight distribution. In general, current predictive models provide explanatory signals, rather than a modeling and validation framework for entity conditioning and using lag distribution as a structured output, so the lag gate + audit approach in this paper is designed for this purpose.

Beyond architecture design, the application of panel machine learning also brings an evaluation problem. Supervised learning metrics are mainly used to evaluate observed predictive error, but they often fail to capture deployment-relevant objectives or the way of using information from the model when prediction alone is not the only goal \cite{lipton2018mythos}. For models that report explanatory structure, this issue is very important, because post-hoc explanations of black-box predictors can be incomplete or unfaithful, and interpretable models should exhibit internal structure that can be directly inspected and audited \cite{rudin2019stop,wang2023counter}. In panel data, the temporal and cross-sectional dependencies further aggravate the complexity of the evaluation problem. Contemporaneous covariates, random unit-time splits, or preprocessing choices that use information unavailable at prediction time can lead to data leakage and exaggerate out-of-sample performance estimates \cite{cerqua2025mlpanel}. As a result, it is not enough to claim that a model has learned meaningful heterogeneous lag structure only based on high predictive accuracy. Our work aims to fill this evaluation gap by treating the learned effective lag as an auditable model-derived summary.

\section{Problem Formulation}\label{sec:problem_formulation}
\subsection{Panel Setup}
Considering a balanced panel time series with $N$ entities and $T$ time steps. For each entity $i \in \{1,\ldots,N\}$ and time $t \in \{1,\ldots,T\}$, a sequential covariate vector $X_{i,t}\in\mathbb{R}^{d_x}$ and a scalar target $Y_{i,t}\in\mathbb{R}$ can be observed. Each entity is associated with a time-invariant proxy vector $p_i\in\mathbb{R}^{M}$ and a static feature vector $s_i\in\mathbb{R}^{d_s}$ when available. This setting follows the standard panel time-series view where temporal dependence and cross-sectional heterogeneity are both central properties of the data \cite{thayasivam2025panelsurvey}. The proxy-as-conditioning view is inspired by absorptive capacity, a kind of capacity to recognize, assimilate and apply external knowledge, which depends on prior related knowledge \cite{cohen1990absorptive}. Its core idea can be abstracted to a more general entity-level (e.g., countries, regions or organizations): entities with different knowledge foundations, institutional conditions or development levels may absorb shocks or inputs with different speeds, and exhibit different delays in future outcomes.

\subsection{Entity-Conditioned Heterogeneous Lag Mining}
Given a maximum lag horizon $K$, the goal is to learn an entity-specific lag-weight distribution:
\[
    \omega_i = (\omega_{i,1}, \ldots, \omega_{i,K}),
    \qquad
    \sum_{k=1}^{K}\omega_{i,k}=1,\quad \omega_{i,k}\ge 0.
\]
where $\omega_{i,k}$ denotes the normalized weight assigned to lag $k$ for entity $i$.

The lag weights aggregate projected past sequential inputs into a per-entity context vector:
\begin{equation}
    c_{i,t} = \sum_{k=1}^{K}\omega_{i,k}\widetilde{X}_{i,t-k}.
\end{equation}
where $\widetilde{X}_{i,t-k}$ denotes the projected sequential representation at lag $k$. Then, the LSTM backbone processes the lag-context stream together with learned entity effects, static features, and optional macro controls to predict the corresponding valid post-warmup target.

The interpretable byproduct is the per-entity effective lag:
\begin{equation}
    k_i^\star = \sum_{k=1}^{K} k \, \omega_{i,k}.
\end{equation}
Because $\omega_i$ only depends on the entity-level conditioning score $z_i$, $k_i^\star$ is constant across time for a fixed entity and can be audited. 

We separate heterogeneous lag mining from standard prediction. Forecasting evaluation focuses on target accuracy, yet lag mining evaluation focuses on non-degeneracy of the learned $k_i^\star$ across entities and whether it is aligned with pre-specified external stratifiers. The learned effective lag $k_i^\star$ is evaluated through the audit protocol described in Section~\ref{sec:audit_protocol}.

\section{Method: AC-GATE}
We propose AC-GATE, an Adaptive-Conditioning Encoder with a Scale-Invariant Lag Gate on top of a recurrent backbone. The model follows the pipeline:
\[
    p_i \rightarrow z_i \rightarrow \omega_i \rightarrow c_{i,t} \rightarrow \hat{Y}_{i,t+1}
\]
This design is motivated by two considerations. First, the absorptive-capacity view abstracted in Section~\ref{sec:problem_formulation} suggests that entity-level prior knowledge can moderate how information is absorbed over time \cite{cohen1990absorptive}. Second, gate- and attention-based sequence models show that input-dependent selection can improve explainability by exposing which parts of a sequence or feature set are used by the model \cite{xue2020gated,lim2021tft}.

\subsection{Adaptive Conditioning Encoder}
For each entity $i$, the adaptive-conditioning encoder maps the entity-level proxy vector $p_i$ to a entity-conditioning score:
\begin{equation}
    z_i = f_{\phi}(p_i),
    \qquad z_i\in\mathbb{R}.
\end{equation}
The proxy vector $p_i$ contains static entity-level covariates that capture persistent, time-invariant heterogeneity across entities. The scalar score $z_i$ is then used to condition the lag gate and is also projected into the initial recurrent state of the LSTM backbone.

A reconstruction head maps the scalar score back to the proxy space:
\begin{equation}
    \widehat{p}_i = g_{\psi}(z_i),
    \qquad \widehat{p}_i\in\mathbb{R}^{M}.
\end{equation}
The corresponding reconstruction term can be used as an auxiliary reconstruction objective. It can serve as a lightweight diagnostic head or an explicit regularizer. In the detached variant, the reconstruction loss only trains the proxy decoder and should not be interpreted as directly regularizing the encoder representation. In the non-detached variant, the reconstruction loss encourages $z_i$ to retain information about the original proxy vector.

\subsection{Scale-Invariant Lag Gate}
Conditioned on the entity-conditioning score $z_i$, the scale-invariant lag gate maps $z_i$ to an entity-specific lag-weight distribution:
\begin{equation}
\begin{aligned}
\omega_{i,k}
&= \mathrm{Softmax}_{k}\!\left(
\frac{g_{\theta}(z_i)_k - \lambda\, k/K}{\tau}
\right),\\
&\qquad k=1,\ldots,K.
\end{aligned}
\end{equation}
where $g_{\theta}(z_i)\in\mathbb{R}^{K}$ outputs entity-specific lag logits, with $g_{\theta}(z_i)_k$ denoting the logit for lag $k$. \textit{Scalar-conditioned} means the logits are produced conditional on the scalar $z_i$. The operator $\mathrm{Softmax}_k(\cdot)$ normalizes these logits over $k=1,\ldots,K$ for each entity $i$.

The relative-position term $\lambda\cdot k/K$ penalizes farther lags on a normalized scale. The normalization by $K$ makes the position bias scale-invariant with respect to the maximum lag horizon. The temperature $\tau$ controls the sharpness of the resulting lag distribution.

\subsection{Backbone and Information-Flow Discipline}
The lag-weighted context $c_{i,t}$ is connected with a learned entity embedding, the optional static feature vector $s_i$, and optional macro-level controls. The sequence is processed by a two-layer LSTM, and the resulting hidden state is mapped to the target prediction.

A key information-flow constraint is that the current sequential observation $X_{i,t}$ is not directly concatenated into the backbone input. Only lagged observations enter through $c_{i,t}$. This prevents a shortcut path where the recurrent backbone could ignore the lag gate and rely on contemporaneous covariates, which is a failure mode closely related to leakage concerns in panel machine learning \cite{cerqua2025mlpanel}.

\begin{figure}[!tbhp]
    \centering
    \includegraphics{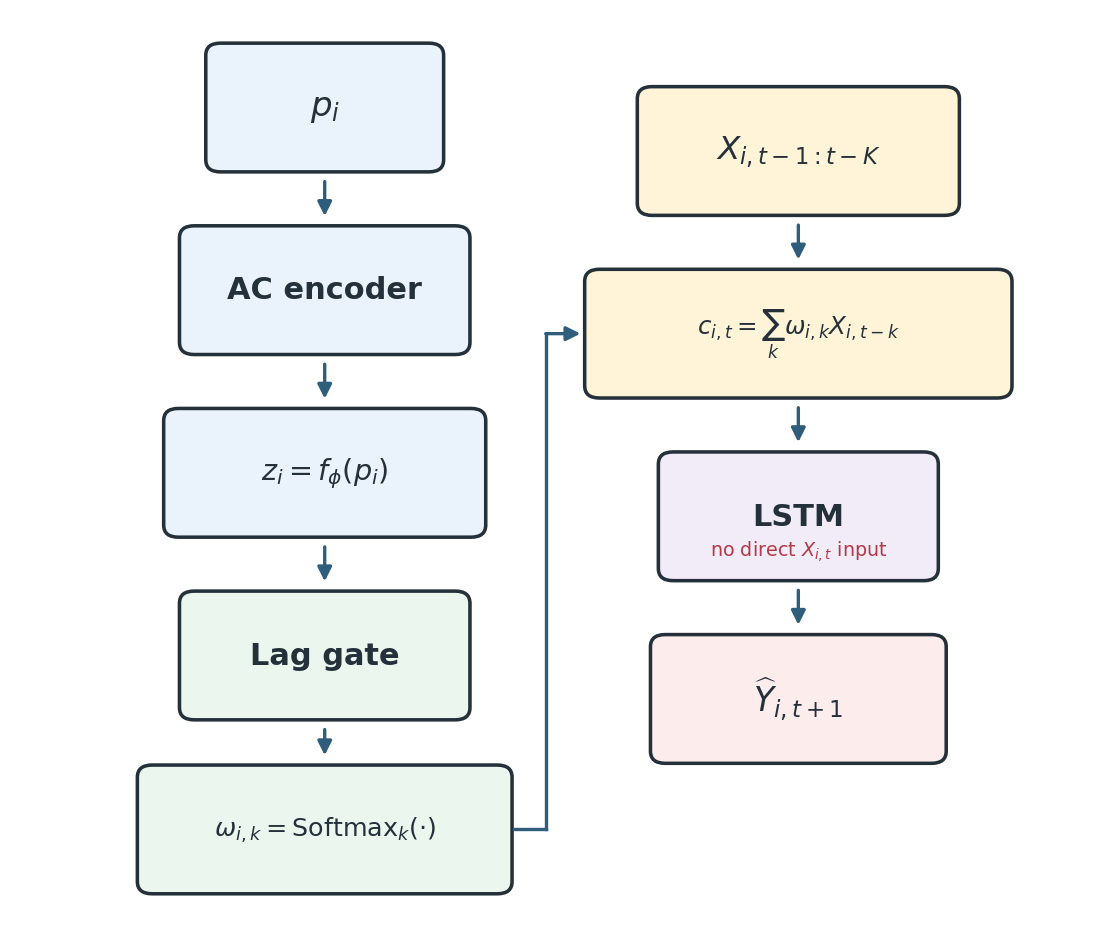}
    \caption{Overview of the AC-GATE architecture.}
    \label{fig:ac_gate_arch}
\end{figure}

\subsection{Training Objective}
Training minimizes the domain-agnostic objective:
\begin{equation}
    \mathcal{L}
    =
    \mathcal{L}_{\mathrm{task}}
    +
    \lambda_r \mathcal{L}_{\mathrm{recon}}.
\end{equation}
The task loss $\mathcal{L}_{\mathrm{task}}$ is the forecasting mean squared error (MSE) between the predicted target and the true target at the corresponding valid post-warmup step. The reconstruction loss $\mathcal{L}_{\mathrm{recon}}$ compares the original proxy vector $p_i$ with the reconstructed proxy vector $\widehat{p}_i$. The coefficient $\lambda_r$ controls the weight of the reconstruction term and is distinct from the lag-position bias coefficient $\lambda$ in the gate.

\subsection{Computational Complexity}
For a panel with $N$ entities, $T$ time steps, $F$ dynamic features, maximum lag $K$, LSTM hidden size $H$, and $L$ recurrent layers, the dominant per-epoch cost of AC-GATE is $O(NT(KF + LH(F+H)))$. The additional entity-conditioned lag gate costs $O(NKH)$ and is computed once per entity, while the lag aggregation costs $O(NTKF)$. So the main computational burden remains the recurrent backbone. At inference, the entity-specific lag distribution $\omega_i$ and effective lag $k_i^\star$ can be precomputed for each entity. The prediction has the same order as the LSTM forward pass plus lag aggregation.

\section{LAG DISCOVERY AND AUDIT PROTOCOL}\label{sec:audit_protocol}
A layered audit protocol is used to evaluate AC-GATE. This reflects the view that the evaluation of predictive models does not focus only on test-set accuracy but also to whether they yield useful, deployment-relevant understanding (e.g., transparency, accountability, or faithful structure) beyond what standard predictive metrics capture \cite{lipton2018mythos,rudin2019stop}. This separation is necessary because forecast accuracy and structured lag discovery are related but different: a model may be weak at real-data forecasting while still producing a non-degenerate and externally structured effective-lag signal. Conversely, a strong forecasting baseline does not automatically imply an interpretable, entity-conditioned lag distribution. Algorithm~\ref{alg:acgate_audit} summarizes the four-layer audit protocol used throughout the experiments.

\begin{algorithm}[!tbhp]
\caption{Lag Discovery and Layered Audit}
\label{alg:acgate_audit}
\Input{$\{X_{i,t},Y_{i,t}\}$; $\{p_i\}$; optional $\{s_i\}$; lag horizon $K$; seeds $\mathcal{S}$; stratifiers $\bm{\xi}$; permutations $B$; threshold $\epsilon$; optional truth $k^{\mathrm{true}}_i$}
\Output{Forecast metrics; $\omega_i$; $k_i^\star$; L0--L3 verdicts}

\ForEach{$r\in\mathcal{S}$}{
Split chronologically\;
Fit normalizers on train only\;
Train AC-GATE by minimizing $\mathcal{L}_{\mathrm{task}}+\lambda_r\mathcal{L}_{\mathrm{recon}}$\;
Select checkpoint on validation\;

\tcp{L0 (forecast):}
Report test MSE/MAE/$R^2$\;

Compute effective lag $k_i^\star\gets\sum_{k=1}^{K}k\,\omega_{i,k}$\;

\tcp{L1 (degeneracy guard):}
Compute $\mathrm{sd}_i(k_i^\star)$\;
\If{$\mathrm{sd}_i(k_i^\star)\leq\epsilon$}{Mark seed degenerate\;\Continue}

\tcp{L2 (structured heterogeneity):}
\ForEach{$\xi^{(j)}\in\bm{\xi}$}{
Compute $\rho_{r,j}\gets\mathrm{Spearman}(k_i^\star,\xi_i^{(j)})$\;
Draw $B$ entity-label permutations of $\xi^{(j)}$ and compute $\rho^{(b)}$\;
Compute $p_{r,j}\gets(1+\#\{b:|\rho^{(b)}|\geq|\rho_{r,j}|\})/(B+1)$\;
}

\tcp{L3 (optional ground truth):}
\eIf{$k^{\mathrm{true}}_i$ available}{Report $\mathrm{MAE}(k^\star,k^{\mathrm{true}})$ and $\mathrm{Spearman}(k^\star,k^{\mathrm{true}})$\;}{Mark n/a\;}
}

Aggregate L0 across seeds\;
Aggregate L2 via $\mathrm{mean}(|\rho|)$, rejection share, and Fisher-combined $p$\;
Run paired seed-level tests vs. baselines/ablations\;
Assemble the L0--L3 verdict matrix\;
\end{algorithm}

For AC-GATE and its lag-gate ablations, $k_i^\star$ is a structural quantity computed from the entity-specific lag distribution $\omega_i$. To baselines without an explicit entity-conditioned lag gate, a diagnostic $k_i^\star$ summary is reported when a comparable temporal-weight or lag-diagnostic output is available. Such baseline $k_i^\star$ values are used only for empirical comparison and are not interpreted as structural lag-gate outputs.

\subsection{Audit Layers and Verdict Rules}
L0 focuses on the predictive calibration of each method. This layer is included to show whether the learned representation remains predicatively calibrated, but it is not the main mechanism-claim layer.

L1 is a degeneracy guard for lag discovery. For each seed, we compute the cross-entity standard deviation of the learned effective lag, $\mathrm{sd}_i(k_i^\star)$. If this value is below a small threshold $\epsilon$, the seed is marked as degenerate and is not used for L2 stratification. This rule prevents constant-lag models (e.g., uniform-lag or collapsed ablations) from being explained as evidence of heterogeneous lag structure.

L2 evaluates whether the learned effective lag is externally structured. For each pre-specified entity stratifier $\xi_i$, we compute the seed-level Spearman correlation between $k_i^\star$ and $\xi_i$, and assess its absolute value against an entity-label permutation null. This makes L2 a sign-robust structural test rather than a directional semantic test: it asks whether learned lags are non-randomly stratified by pre-specified entity-level variables, without requiring the sign of the association to be stable across seeds. L2 evidence is summarized by the mean $|\rho|$ across seeds, the fraction of seeds with permutation $p<0.05$, and a Fisher-combined permutation $p$-value, as reported in Table~\ref{tab:l2_sign_robust_quantities}. Because stratifiers are constructed from pre-test historical information, the test evaluates alignment with pre-existing entity characteristics rather than future test outcomes.

\begin{table}[!tbhp]
\centering
\caption{Sign-robust L2 summary statistics.}
\label{tab:l2_sign_robust_quantities}
\setlength{\tabcolsep}{5pt}
\renewcommand{\arraystretch}{1.1}
\begin{tabularx}{\columnwidth}{@{}l l@{}}
\toprule
Quantity & Evidence role \\
\midrule
$\mathrm{mean}(|\rho|)$ & Effect-size magnitude across random seeds \\
$\Pr\bigl(p_{\mathrm{perm}}<0.05\bigr)$ & Seed-level stability across random seeds \\
$p_{\mathrm{Fisher}}$ & Aggregated evidence across random seeds \\
\bottomrule
\end{tabularx}
\par\smallskip
{\footnotesize\raggedright\textit{Notes.} L2 uses absolute Spearman alignment because the sign of a learned latent lag score can vary across seeds. Permutation tests shuffle entity labels of $\xi_i$ while holding $k_i^\star$ fixed.\par}
\end{table}

L3 is only evaluated when ground-truth mechanism information is available. In the synthetic data, we compare $k_i^\star$ with the true lag-generating target via MAE and Spearman correlation. In the real-world panels, L3 is marked as not applicable because no supervised mechanism label is observed. Accordingly, real-data evidence is restricted to L1/L2 structured effective-lag summaries rather than supervised ground-truth recovery.

\section{Experiments}
\subsection{Experimental Setup}

AC-GATE is evaluated on three domains with distinct evidential roles:
\begin{itemize}
\item \textbf{Synthetic:}
The synthetic setting provides ground-truth heterogeneous lags for mechanism recovery. See Appendix~\ref{app:synthetic_dgp} for the script that generates the balanced panel time series used here in detail. 
\item \textbf{PWT 11.0:}
 Economic panel data with the target variable $ctfp$ and an effective-labor-aware feature bundle, where 1980--2007 for training, 2008--2013 for validation, and 2014--2023 for testing. PWT provides internationally comparable measures of output, capital, human capital, labor input and productivity, which make it suitable for cross-country productivity analysis \cite{feenstra2015pwt}.
\item \textbf{OWID Energy \& WGI:}
We combine OWID energy data with WGI; the target is \texttt{co2\_per\_unit\_energy}, where 1996--2011 for training, 2012--2017 for validation, and 2018--2023 for testing. The WGI data provide governance-related entity stratifiers across six broad governance dimensions \cite{kaufmann2011wgi}.
\end{itemize}

In the economic domain, prior work emphasizes cross-sectional parameter heterogeneity. Regime-dependent relationships, motivating designs that avoid relying solely on pooled mean effects \cite{canay2011sim,hansen1999thr,pesaran1995pmg}. And in the energy literature, the relationship between energy transition and emissions is also documented to vary across countries and quantiles \cite{akram2020energy,mirziyoyeva2022re}.

To support reproducibility and ensure controlled comparison, we summarize the baseline/ablation design choices that define the controlled comparison in Table~\ref{tab:fair_baselines}. Baselines include Plain LSTM, TFT \cite{lim2021tft}, GA-Net \cite{xue2020gated}, and Grouped ARDL\cite{pesaran1995pmg,babii2021mlpanel}. Structural ablations are No-AC-Encoder ($z_i$ fixed), Uniform-Lag ($\omega_{i,k}=1/K$), and No-Recon-Reg ($\lambda_r=0$). Details of experimental parameters and data preprocessing are provided in the appendix~\ref{app:repro}. In addition, we also include a negative-control check via proxy permutation to diagnose spurious predictability, which is shown in Table~\ref{tab:proxy_shuffle_setup}.

\begin{table}[!tbhp]
\caption{controlled comparison design.}
\label{tab:fair_baselines}
\centering
\scriptsize
\setlength{\tabcolsep}{3pt}
\renewcommand{\arraystretch}{1.03}
\begin{tabular}{p{0.22\columnwidth} p{0.72\columnwidth}}
\toprule
Method & Difference (all share the same splits, $K$, and protocol) \\
\midrule
AC-GATE & Full model (AC encoder + lag gate + recon.). \\
Plain LSTM & No AC encoder; no lag gate (matched backbone). \\
TFT & Transformer with static covariates and temporal attention. \\
GA-Net & Gated-attention baseline. \\
No-AC-Encoder & Fix $z_i$ (no entity-conditioned lag). \\
Uniform-Lag & Uniform $\omega_{i,k}$ (no learned lag heterogeneity). \\
No-Recon-Reg & Set $\lambda_r=0$ (no recon.). \\
Grouped ARDL & Econ./Energy only: grouped OLS distributed-lag baseline. \\
\bottomrule
\end{tabular}
\par\smallskip
{\footnotesize\raggedright\textit{Notes.} Plain LSTM has the same backbone capacity as AC-GATE, but removes the adaptive-conditioning encoder and lag gate. Grouped ARDL uses the same feature sets and the same $K$-lag block, and grouping only uses training-window information.\par}
\end{table}

\begin{table}[!tbhp]
\centering
\caption{Proxy-Shuffle Negative Control Setup}
\label{tab:proxy_shuffle_setup}
\footnotesize
\setlength{\tabcolsep}{3pt}
\renewcommand{\arraystretch}{1.05}
\setlength{\emergencystretch}{1em} 
\begin{tabularx}{\columnwidth}{@{}>{\RaggedRight\arraybackslash}p{0.23\columnwidth} >{\RaggedRight\arraybackslash}X@{}}
\toprule
Aspect & Setting \\
\midrule
Purpose & Test if L2 needs correct proxy--entity match. \\
Control type & Information negative control (not ablation). \\
Shuffle & Permute proxy vectors across entities. \\
Held fixed & Data/IDs; splits/strata; architecture; hyperparameters. \\
Permutation rule & Per seed, draw one non-identity proxy permutation. \\
Training & Retrain AC-GATE for 20 seeds (same protocol). \\
Evaluation & Recompute stratified L2 $k^*$ audit (original stratifiers). \\
Primary metric & Mean $|\rho|$ drop; reject rate; Fisher-combined $p$. \\
Interpretation & Collapse implies proxy--entity dependence. \\
\bottomrule
\end{tabularx}
\end{table}

\subsection{Paired significance tests}
All models are evaluated via 20 random seeds. For paired model comparisons, we use a paired Wilcoxon test on seed-level metrics, which compares AC-GATE against each baseline or ablation model under the same seeds. This scheme detects systematic deviations of paired differences from zero while controlling for training variability across seeds.

For the synthetic experiments, the main test metric is $k^\star$ MAE because the ground-truth effective lag is known. For the real-data experiments, we do not use Wilcoxon tests to validate real-data lag-audit claims. Paired tests are applied only to forecast-layer metrics like $R^2$. Stratifier alignment is evaluated separately through entity-level permutation tests in each seed and Fisher aggregation across seeds. Detailed paired-test results are reported in Appendix~\ref{app:wilcoxon}.

\subsection{Experimental results and analysis}

\begin{figure}[!!tbhp]
    \centering    
    \includegraphics[width=\columnwidth]{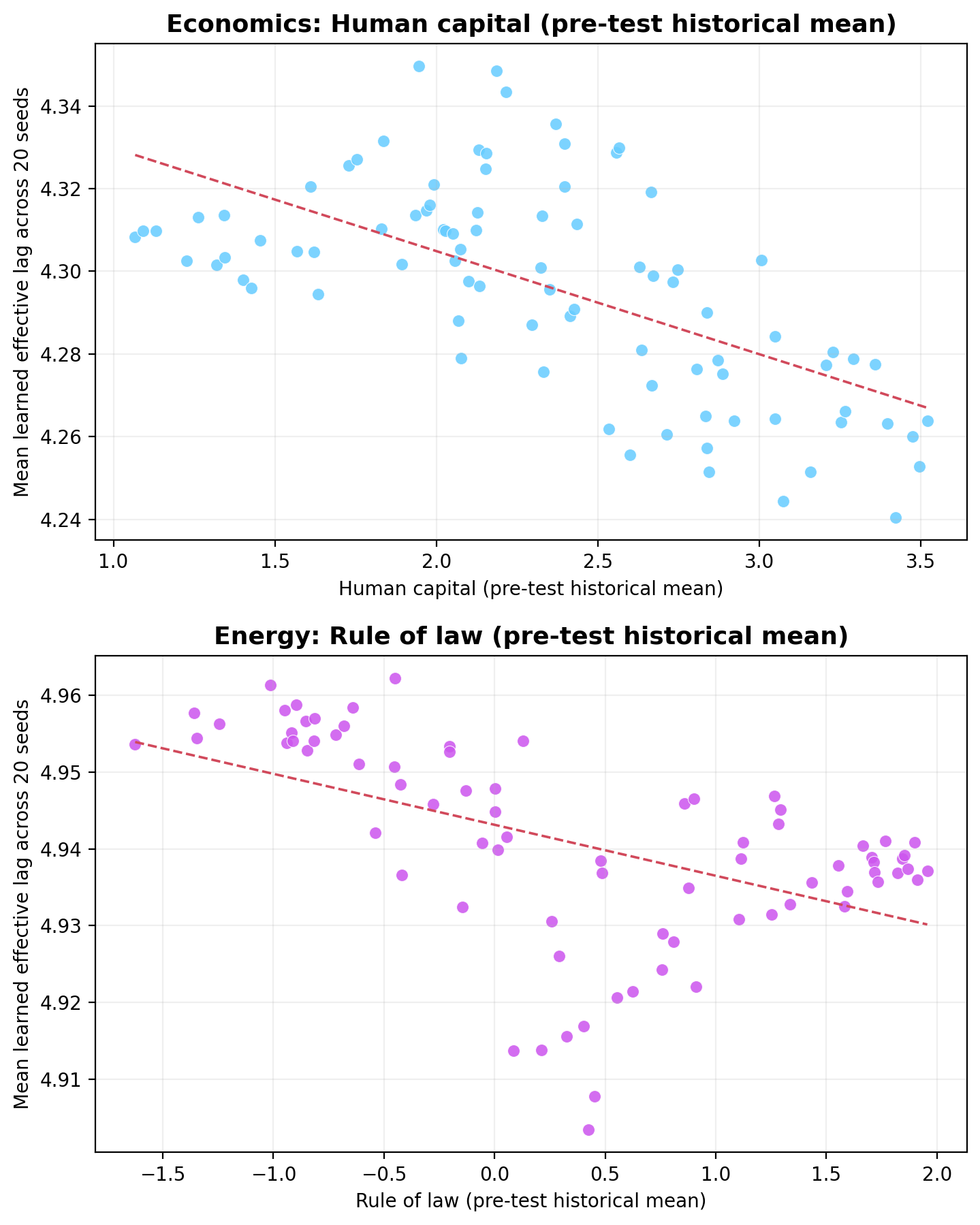}
    \caption{Relationship between entity-level stratifiers and learned effective lag.}
    {\footnotesize\raggedright\textit{Notes.} Only use pre-test historical information. Per-entity $k^\star$ is the mean over 20 seeds; cross-seed averages are used for visualization, while the inferential L2 claim uses sign-robust statistics (see Section~\ref{sec:audit_protocol}). \par}
    \label{fig:case_study_panel}
\end{figure}

\begin{figure*}[!t]
    \centering
    \includegraphics[width=\textwidth]{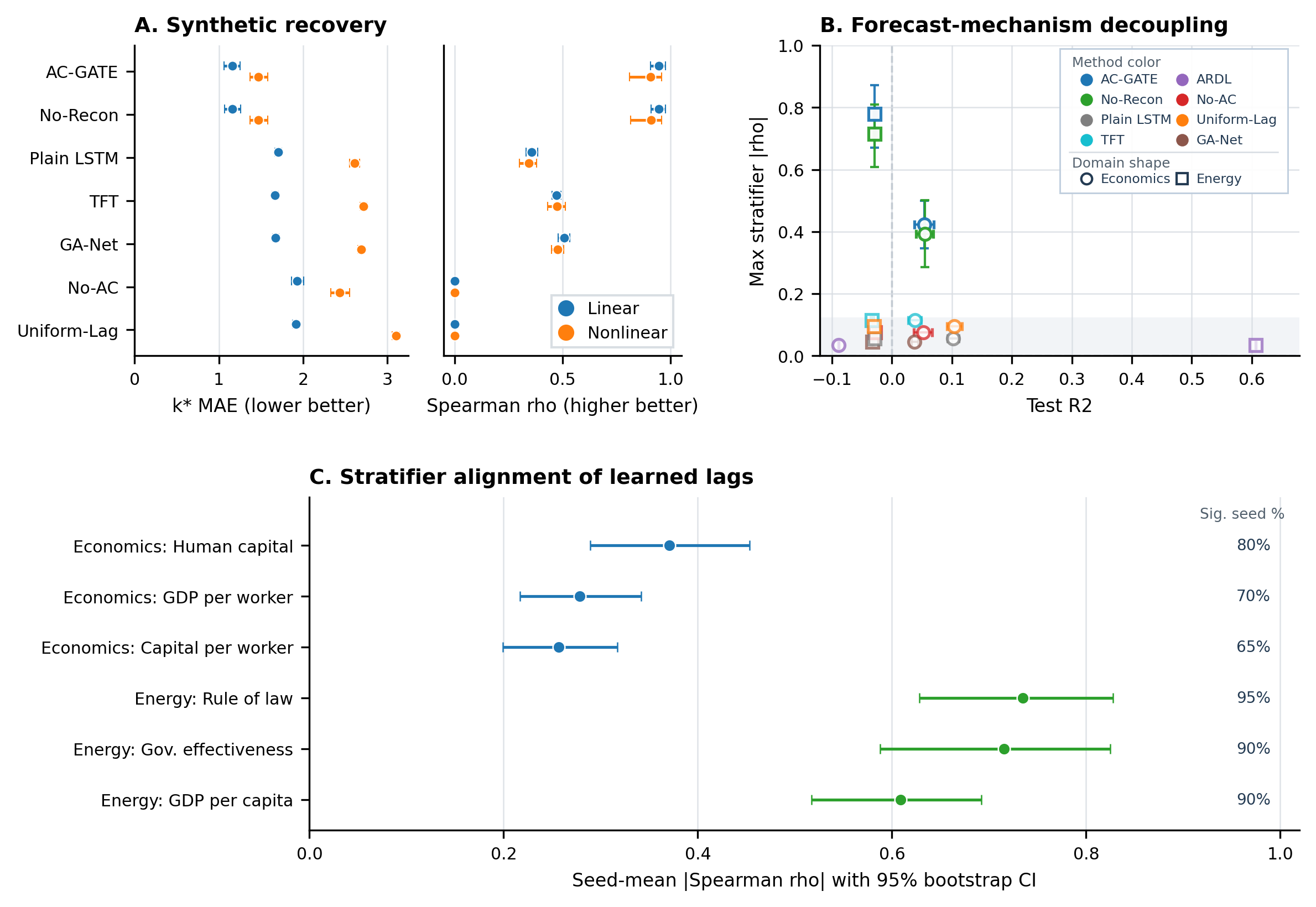}
    \caption{Evidence for AC-GATE recovery, forecast-mechanism decoupling, and stratifier alignment.}
    \label{fig:compact_evidence_panel}
    {\footnotesize\raggedright\textit{Notes.} Panel A summarizes synthetic mechanism recovery under linear and nonlinear settings. Panel B contrasts predictive performance with the strength of learned stratifier association in real-data tasks. Panel C reports AC-GATE stratifier alignment across economics and energy variables, with the right column showing the percentage of seeds whose permutation-test p-value is below 0.05.\par}
\end{figure*}

In Fig.~\ref{fig:case_study_panel}, AC-GATE's learned entity-level effective lag \(k_i^\star\) is negatively associated with external stratifiers: human capital in the economics panel (Spearman \(\rho=-0.63\), \(p\approx 3\times 10^{-11}\)) and rule of law in the energy panel (\(\rho=-0.61\), \(p\approx 3\times 10^{-9}\)). This descriptive pattern indicates that AC-GATE assigns shorter effective memory to entities with stronger institutional or capacity-related stratifiers, while the formal L2 claim remains sign-robust external alignment. The direction is consistent with absorptive- and institutional-capacity arguments: stronger human capital and governance may reduce learning and implementation frictions in technology adoption, while rule-of-law and policy-capacity conditions may support more structured renewable-energy transition responses \cite{cohen1990absorptive,schweikl2020lessons,kaufmann2011wgi,mirziyoyeva2022re,akram2020energy}.

Fig.~\ref{fig:compact_evidence_panel} summarizes the evidence for AC-GATE recovery. The synthetic experiments test recovery of known heterogeneous lag structure. AC-GATE achieves strong rank alignment between learned and true effective lags (Spearman $\rho \approx 0.945$ linear; $0.907$ nonlinear), while No-AC-Encoder and Uniform-Lag show near-zero alignment due to structurally collapsed entity-level lag variation.

In the economics domain, the learned effective-lag summary is non-degenerate and stratified. AC-GATE maintains entity-level lag variation ($k^\star$ std. $\approx 0.167$) and lag-gate sensitivity, while No-AC-Encoder and Uniform-Lag collapse the structured lag signal. Moreover, learned effective lags of AC-GATE align with pre-specified economics stratifiers: human capital shows the strongest association (mean $|\rho| \approx 0.371$), followed by GDP per worker ($0.278$) and capital per worker ($0.257$), with 65--80\% of seeds significant under the permutation test.

Energy-domain diagnostics provide supportive evidence: AC-GATE and No-Recon retain non-degenerate entity-level lag variation, whereas No-AC-Encoder and Uniform-Lag collapse it. AC-GATE effective lags exhibit large sign-robust associations with rule of law, government effectiveness, and GDP per capita (mean $|\rho| \approx 0.735$, $0.716$, and $0.609$), with $90$--$95\%$ seed-level permutation significance.

\begin{figure}[!tbhp]
  \centering
  \includegraphics[width=\columnwidth]{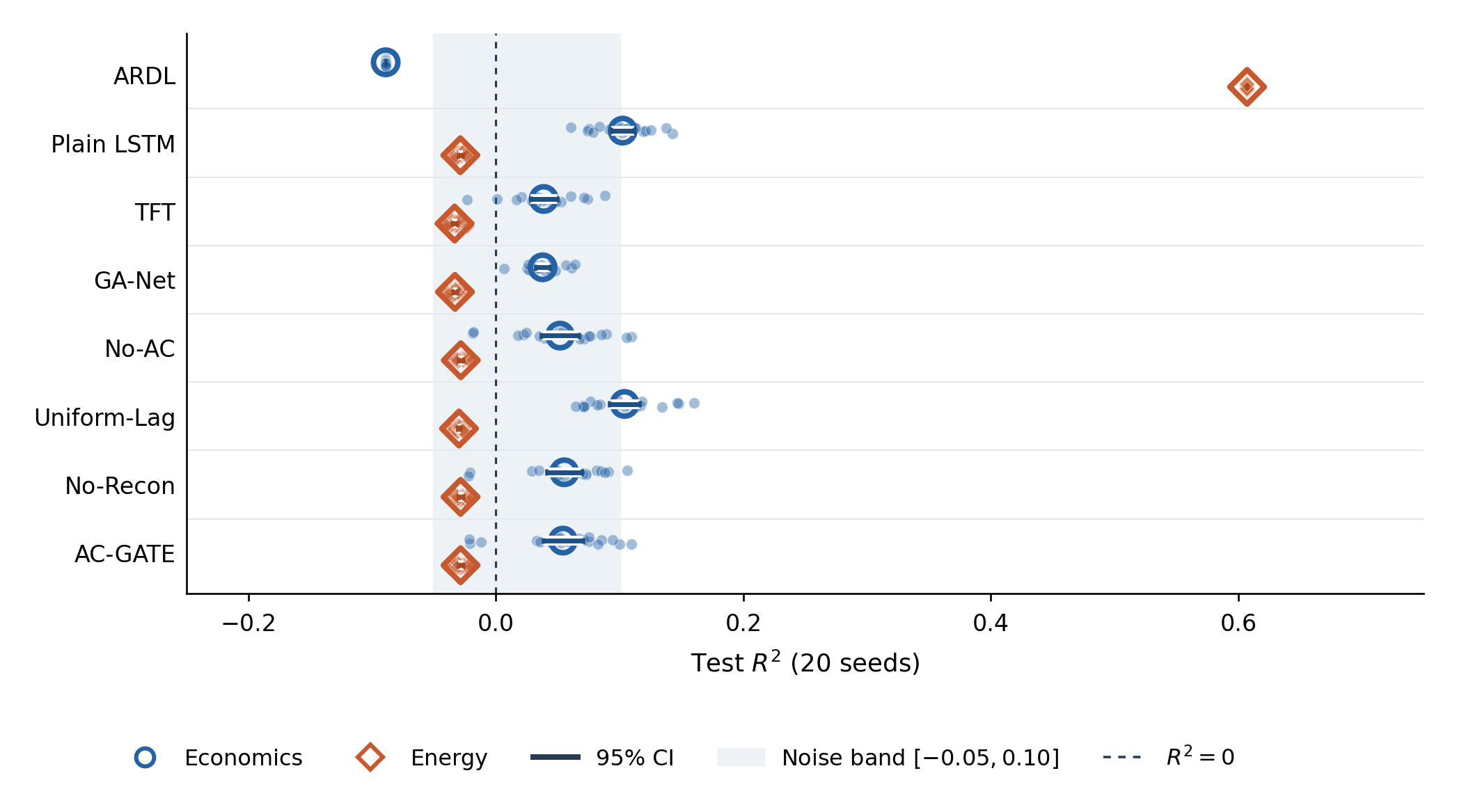}
  \caption{Forecast $R^2$ on the two real-world panels.}
  {\footnotesize\raggedright\textit{Notes.} All deep baselines (incl. AC-GATE and ablations) fall in the near-zero band $[-0.05, 0.10]$. Grouped ARDL flips sign across domains ($-0.089$ in Economics; $+0.607$ in Energy), so forecast $R^2$ alone gives unstable rankings; interpret it alongside L1/L2 audit evidence (Sec.~\ref{sec:audit_protocol}).\par}
  \label{fig:r2_floor}
\end{figure}

Fig.~\ref{fig:r2_floor} shows that the forecast \(R^2\) values in real-data panels cluster around a narrow band near zero for AC-GATE and all deep learning baselines, except for the Grouped ARDL in the energy domain which achieves a significantly higher \(R^2\). These results show that AC-GATE is not a forecasting-dominant model on the real panels; the real-data contribution is therefore evaluated through L1/L2 lag-audit evidence rather than L0 superiority.

As shown in Table~\ref{tab:proxy_shuffle_compact}, the proxy-shuffle negative control breaks the proxy--entity correspondence in real-data panels. In this control, L2 alignment fails and predictive performance remains nearly unchanged. This indicates that the stratified \(k^\star\) in real domains is sensitive to the correct proxy--entity pairing, not explained by generic model capacity or unchanged forecasting performance alone.

In summary, the real-data experiments show that the strongest evidence for AC-GATE lies in structured and non-degenerate effective-lag summaries. Based on the experimental results and audit protocol, Table~\ref{tab:verdict_matrix} reports the verdict matrix.

\begin{table}[!tbhp]
\centering
\caption{Proxy-shuffle negative control on real-data L2 alignment.}
\label{tab:proxy_shuffle_compact}
\scriptsize
\setlength{\tabcolsep}{2pt}
\renewcommand{\arraystretch}{1.05}
\begin{tabular}{llccccc}
\toprule
Dom. & Model & Mean $|\rho|$ & $\mathrm{sd}(k^\star)$ & Frac. $p<5\%$ & Fisher $p$ & Test $R^2$ \\
\midrule
Econ. & AC-GATE & 0.302 & 0.168 & 0.72 & $10^{-46}$--$10^{-24}$ & 0.054 \\
Econ. & Proxy-shuf. & 0.058 & 0.184 & 0.00 & 0.94--1.00 & 0.055 \\
Ener. & AC-GATE & 0.686 & 0.109 & 0.92 & $10^{-79}$--$10^{-77}$ & -0.029 \\
Ener. & Proxy-shuf. & 0.088 & 0.073 & 0.07 & 0.39--0.75 & -0.028 \\
\bottomrule
\end{tabular}
\vspace{1mm}
\begin{flushleft}
\footnotesize \textit{Note}: Economics stratifiers are human capital, capital per worker, and GDP per worker. Energy stratifiers are government effectiveness, GDP per capita, and rule of law. Proxy shuffling preserves model architecture and training protocol but breaks the entity--proxy correspondence.
\end{flushleft}
\end{table}

\begin{table}[!tbhp]
\centering
\caption{Verdict matrix.}
\label{tab:verdict_matrix}
\renewcommand{\arraystretch}{1.25}
\setlength{\tabcolsep}{6pt}
\begin{tabularx}{\columnwidth}{l *{4}{>{\centering\arraybackslash}X}}
\toprule
Domain & L0 & L1 & L2 & L3 \\
\midrule
Synthetic & \nc & \yes & \yes & \yes \\
Economics & \nc & \yes & \yes & \NA \\
Energy & \nc & \yes & \yes & \NA \\
\bottomrule
\end{tabularx}
\par\smallskip
{\footnotesize\raggedright Green denotes supported, amber not certified, red ruled out, and gray not claimed.\par}
\end{table}

\section{Discussion}
Table~\ref{tab:verdict_matrix} shows that AC-GATE is supported on L1/L2 in all domains. The experiments show that the learned $k^\star$ should be regarded as a model-derived effective lag summary, not as a causal delay estimate. In synthetic, $k^\star$ can be evaluated directly against the known data-generating lag. In real panels, the analogous question is still testable: whether the learned entity-level lags are non-randomly stratified by pre-specified external indicators. These findings support AC-GATE as a structured lag-audit tool for characterizing heterogeneous temporal response patterns in observational panels.

The ablation results clarify where the lag-audit evidence comes from. Removing the AC encoder or replacing the lag gate with a uniform lag distribution eliminates entity-level lag variation, thereby making stratifier alignment impossible. By contrast, removing the reconstruction regularizer hardly changes synthetic lag recovery, which indicates that the reconstruction term is an auxiliary stabilizer, rather than the source of synthetic recovery or real-data alignment. As a result, the essential inductive bias is the combination of entity conditioning and explicit lag gating: the model must both represent entity-level context and convert that context into a non-uniform lag distribution. Taken together, the layered audit and the ablations identify a stronger claim than forecasting gain: AC-GATE provides a verifiable way to expose entity-conditioned lag heterogeneity as a structured model output.

AC-GATE is not universally superior in forecasting. In the synthetic with known heterogeneous lag structure, AC-GATE has high rank consistency in recovering the true effective lag. However, in real-world data, the predictive performance of AC-GATE does not show an advantage. This distinction is crucial for understanding the limitations of AC-GATE: it is designed to reveal non-degenerate, entity-conditioned lag structure in panel data. Therefore, the real-data results support structured lag-audit evidence and do not claim forecasting superiority. This separates the present work from causal panel learning methods that explicitly target heterogeneous treatment effects and counterfactual recovery \cite{zhou2025codeal}.

\section{Conclusion \& Future Work}
This paper contributes a framework for auditing entity-conditioned lag heterogeneity in panel time series. AC-GATE operationalizes this framework by using observable entity-level proxies to produce explicit lag-gate distributions and entity-level effective-lag summaries. Across synthetic and real-world panels, the evaluation shows that these summaries can recover known lag structure when ground truth is available and remain non-degenerate and externally structured in applied country-level settings. The contribution is therefore not a new forecasting benchmark, but a reproducible lag-audit design for turning heterogeneous temporal response patterns into inspectable model outputs. In this sense, the paper shifts lag analysis in panel learning from an informal interpretation layer to an explicit audit target. The output of interest is not only a prediction, but also an inspectable entity-level lag summary whose heterogeneity can be checked for collapse, external structure, and synthetic recoverability under a common protocol.

Several directions can be retained for future work. First, further exploring the performance of AC-GATE on different types of panel data, and its applicability to different prediction tasks. Second, trying to introduce more complex entity conditioning encoders or more flexible lag gating mechanisms to enhance the ability of the model to capture complex lag structures. Finally, combining domain knowledge to design more targeted external stratifiers, and validating the relationship between the lag structure discovered by AC-GATE and domain-validated mechanisms in more empirical backgrounds, to further enhance its explanatory power and practical value.

\section*{Acknowledgments}
The authors used chatGPT and GitHub Copilot to assist with polishing and coding. All experimental design, implementation, analysis, and conclusions are the authors' own.

\FloatBarrier
\clearpage
\appendices

\section{Paired Wilcoxon Significance Tests}\label{app:wilcoxon}
\subsection{Synthetic: \(k^\star\) MAE \& Forecast Task Loss}
\begin{table}[!tbhp]
\centering
\caption{Paired Wilcoxon test of \(k^\star\) MAE on the synthetic.}
\label{tab:app_wilcoxon_synth_kstar}
\footnotesize
\setlength{\tabcolsep}{4pt}
\noindent\begin{tabular*}{\columnwidth}{@{\extracolsep{\fill}} l l r r r r}
\toprule
Scenario & Method & Mean diff. & Median diff. & $W$ & $p$ \\
\midrule
linear    & No AC Encoder       & $+0.772$ & $+0.785$ & $0$  & $1.9{\times}10^{-6}$ \\
linear    & No Recon Reg.       & $+0.000$ & $+0.000$ & $86$ & $0.498$ \\
linear    & Plain LSTM          & $+0.548$ & $+0.588$ & $0$  & $1.9{\times}10^{-6}$ \\
linear    & TFT                 & $+0.508$ & $+0.545$ & $0$  & $1.9{\times}10^{-6}$ \\
linear    & GA-Net              & $+0.515$ & $+0.526$ & $0$  & $1.9{\times}10^{-6}$ \\
linear    & Uniform Lag         & $+0.754$ & $+0.799$ & $0$  & $1.9{\times}10^{-6}$ \\
nonlinear & No AC Encoder       & $+0.968$ & $+0.972$ & $0$  & $1.9{\times}10^{-6}$ \\
nonlinear & No Recon Reg.       & $-0.000$ & $-0.000$ & $80$ & $0.368$ \\
nonlinear & Plain LSTM          & $+1.144$ & $+1.122$ & $0$  & $1.9{\times}10^{-6}$ \\
nonlinear & TFT                 & $+1.251$ & $+1.264$ & $0$  & $1.9{\times}10^{-6}$ \\
nonlinear & GA-Net              & $+1.222$ & $+1.226$ & $0$  & $1.9{\times}10^{-6}$ \\
nonlinear & Uniform Lag         & $+1.635$ & $+1.669$ & $0$  & $1.9{\times}10^{-6}$ \\
\bottomrule
\end{tabular*}
\par\smallskip
{\footnotesize\raggedright\textit{Notes.} Mean/median diff.\ $=$ method $-$ AC-GATE.(reference = AC-GATE, 20 seeds)\par}
\end{table}

\begin{table}[!tbhp]
\centering
\caption{Paired Wilcoxon test of forecast task loss on the synthetic.}
\label{tab:app_wilcoxon_synth_taskloss}
\footnotesize
\setlength{\tabcolsep}{4pt}
\noindent\begin{tabular*}{\columnwidth}{@{\extracolsep{\fill}} l l r r r r}
\toprule
Scenario & Method & Mean diff. & Median diff. & $W$ & $p$ \\
\midrule
linear    & No AC Encoder       & $+0.0267$ & $+0.0267$ & $0$  & $1.9{\times}10^{-6}$ \\
linear    & No Recon Reg.       & $+0.0001$ & $-0.0000$ & $98$ & $0.812$ \\
linear    & Plain LSTM          & $+0.0333$ & $+0.0338$ & $0$  & $1.9{\times}10^{-6}$ \\
linear    & TFT                 & $-0.0043$ & $-0.0032$ & $15$ & $2.6{\times}10^{-4}$ \\
linear    & GA-Net              & $-0.0037$ & $-0.0065$ & $59$ & $0.090$ \\
linear    & Uniform Lag         & $+0.0333$ & $+0.0333$ & $0$  & $1.9{\times}10^{-6}$ \\
nonlinear & No AC Encoder       & $+0.0282$ & $+0.0352$ & $0$  & $1.9{\times}10^{-6}$ \\
nonlinear & No Recon Reg.       & $+0.0000$ & $-0.0000$ & $85$ & $0.475$ \\
nonlinear & Plain LSTM          & $+0.0549$ & $+0.0543$ & $0$  & $1.9{\times}10^{-6}$ \\
nonlinear & TFT                 & $-0.0022$ & $-0.0007$ & $83$ & $0.430$ \\
nonlinear & GA-Net              & $+0.0029$ & $+0.0044$ & $33$ & $5.6{\times}10^{-3}$ \\
nonlinear & Uniform Lag         & $+0.0560$ & $+0.0585$ & $0$  & $1.9{\times}10^{-6}$ \\
\bottomrule
\end{tabular*}
\par\smallskip
{\footnotesize\raggedright\textit{Notes.} Mean/median diff.\ $=$ method $-$ AC-GATE. (reference = AC-GATE, 20 seeds; lower is better)\par}
\end{table}

\subsection{Real Domains: Test \(R^2\)}
\begin{table}[!tbhp]
\centering
\caption{Paired Wilcoxon test of test \(R^2\) on the real-data panels.}
\label{tab:app_wilcoxon_real_r2}
\footnotesize
\setlength{\tabcolsep}{4pt}
\noindent\begin{tabular*}{\columnwidth}{@{\extracolsep{\fill}} l l r r r r}
\toprule
Domain & Method & Mean diff. & Median diff. & $W$ & $p$ \\
\midrule
Economics & Grouped ARDL  & $-0.1433$ & $-0.1447$ & $0$  & $1.9{\times}10^{-6}$ \\
Economics & No AC Encoder & $-0.0023$ & $-0.0007$ & $68$ & $0.177$ \\
Economics & No Recon Reg. & $+0.0011$ & $-0.0009$ & $78$ & $0.330$ \\
Economics & Plain LSTM    & $+0.0483$ & $+0.0413$ & $9$  & $6.3{\times}10^{-5}$ \\
Economics & TFT           & $-0.0155$ & $-0.0240$ & $64$ & $0.133$ \\
Economics & GA-Net        & $-0.0164$ & $-0.0258$ & $56$ & $0.070$ \\
Economics & Uniform Lag   & $+0.0498$ & $+0.0410$ & $0$  & $1.9{\times}10^{-6}$ \\
\midrule
Energy    & Grouped ARDL  & $+0.6356$ & $+0.6352$ & $0$  & $1.9{\times}10^{-6}$ \\
Energy    & No AC Encoder & $+0.0002$ & $+0.0001$ & $97$ & $0.784$ \\
Energy    & No Recon Reg. & $+0.0000$ & $+0.0000$ & $24$ & $1.4{\times}10^{-3}$ \\
Energy    & Plain LSTM    & $-0.0002$ & $-0.0020$ & $95$ & $0.729$ \\
Energy    & TFT           & $-0.0048$ & $-0.0046$ & $36$ & $8.3{\times}10^{-3}$ \\
Energy    & GA-Net        & $-0.0046$ & $-0.0055$ & $35$ & $7.3{\times}10^{-3}$ \\
Energy    & Uniform Lag   & $-0.0013$ & $+0.0002$ & $86$ & $0.498$ \\
\bottomrule
\end{tabular*}
\par\smallskip
{\footnotesize\raggedright\textit{Notes.} Mean/median diff.\ $=$ method $-$ AC-GATE. (reference = AC-GATE, 20 seeds; higher is better)\par}
\end{table}

\section{Full Multi-seed and Audit Summaries}\label{app:audit_full}

\FloatBarrier
\subsection{Synthetic L0/L3 Summary}
\par\medskip
\begin{table}[!tbhp]
\caption{Synthetic 20-seed summary.}
\label{tab:app_synth_summary}
\scriptsize
\setlength{\tabcolsep}{3pt}
\noindent\begin{tabular*}{\columnwidth}{@{\extracolsep{\fill}} l l r r r}
\toprule
Scenario & Method & Task loss & \(k^\star\) MAE & Spearman \(\rho\) \\
\midrule
linear & AC-GATE              & $0.0362\,(0.0044)$ & $1.159\,(0.227)$ & $\;\;0.945\,(0.078)$ \\
linear & No Recon Reg.        & $0.0363\,(0.0043)$ & $1.159\,(0.227)$ & $\;\;0.945\,(0.078)$ \\
linear & Plain LSTM           & $0.0695\,(0.0029)$ & $1.707\,(0.091)$ & $\;\;0.356\,(0.066)$ \\
linear & TFT                  & $0.0319\,(0.0025)$ & $1.668\,(0.087)$ & $\;\;0.472\,(0.047)$ \\
linear & GA-Net               & $0.0325\,(0.0069)$ & $1.674\,(0.087)$ & $\;\;0.508\,(0.065)$ \\
linear & No AC Encoder        & $0.0629\,(0.0054)$ & $1.931\,(0.162)$ & $\;\;0.000\,(0.000)$ \\
linear & Uniform Lag          & $0.0695\,(0.0027)$ & $1.913\,(0.088)$ & $\;\;0.000\,(0.000)$ \\
\midrule
nonlin.& AC-GATE              & $0.0378\,(0.0065)$ & $1.467\,(0.248)$ & $\;\;0.907\,(0.212)$ \\
nonlin.& No Recon Reg.        & $0.0378\,(0.0067)$ & $1.467\,(0.248)$ & $\;\;0.909\,(0.205)$ \\
nonlin.& Plain LSTM           & $0.0927\,(0.0061)$ & $2.611\,(0.138)$ & $\;\;0.344\,(0.092)$ \\
nonlin.& TFT                  & $0.0356\,(0.0047)$ & $2.718\,(0.094)$ & $\;\;0.474\,(0.095)$ \\
nonlin.& GA-Net               & $0.0408\,(0.0029)$ & $2.689\,(0.091)$ & $\;\;0.477\,(0.067)$ \\
nonlin.& No AC Encoder        & $0.0661\,(0.0184)$ & $2.435\,(0.261)$ & $\;\;0.000\,(0.000)$ \\
nonlin.& Uniform Lag          & $0.0938\,(0.0081)$ & $3.102\,(0.098)$ & $\;\;0.000\,(0.000)$ \\
\bottomrule
\end{tabular*}
\par\smallskip
{\footnotesize\raggedright\textit{Notes.} Cells report mean (sd) over 20 seeds.Task loss is the forecasting MSE;
\(k^\star\) MAE and Spearman \(\rho(k^\star,k^\star_{\mathrm{true}})\)
are L3 mechanism-recovery indicators.\par}
\end{table}

\FloatBarrier
\subsection{Real-Domain Compact Summaries (L0 \& L1)}
\par\medskip
\begin{table}[!tbhp]
\centering
\caption{Economics 20-seed compact summary. }
\label{tab:app_econ_compact}
\scriptsize
\setlength{\tabcolsep}{3pt}
\noindent\begin{tabular*}{\columnwidth}{@{\extracolsep{\fill}} l r r r r}
\toprule
Method & Test \(R^2\) & \(k^\star_{\sigma}\) & \(\omega\)-entropy & Gate sensitivity \\
\midrule
AC-GATE        & $\;\;0.0541\,(0.0373)$ & $0.167$ & $1.853$ & $0.499$ \\
No Recon Reg.  & $\;\;0.0553\,(0.0329)$ & $0.109$ & $1.844$ & $0.339$ \\
No AC Encoder  & $\;\;0.0518\,(0.0354)$ & $0.000$ & $1.839$ & $0.000$ \\
Uniform Lag    & $\;\;0.1040\,(0.0280)$ & $0.000$ & $2.303$ & $0.735$ \\
Plain LSTM     & $\;\;0.1024\,(0.0214)$ & $1.894$ & $1.873$ & --- \\
TFT            & $\;\;0.0386\,(0.0251)$ & $1.666$ & $1.968$ & --- \\
GA-Net         & $\;\;0.0377\,(0.0133)$ & $1.769$ & $1.912$ & --- \\
Grouped ARDL   & $-0.0891\,(0.0000)$    & ---     & ---     & --- \\
\bottomrule
\end{tabular*}

\par\smallskip
{\footnotesize\raggedright\textit{Notes.} \(k^\star_{\sigma}\) is the mean of the cross-entity standard deviation of \(k^\star\) per seed (L1); a value close to zero indicates a degenerate lag distribution. For AC-GATE variants, $k^\star_\sigma$ is computed from the learned lag-gate distribution. For Plain LSTM, TFT, and GA-Net, $k^\star_\sigma$ is a diagnostic lag-summary statistic computed for comparison and is not a structural gate-derived effective lag.\par}
\end{table}

\begin{table}[!tbhp]
\caption{Energy 20-seed compact summary.}
\label{tab:app_energy_compact}
\scriptsize
\setlength{\tabcolsep}{3pt}
\noindent\begin{tabular*}{\columnwidth}{@{\extracolsep{\fill}} l r r r r}
\toprule
Method & Test \(R^2\) & \(k^\star_{\sigma}\) & \(\omega\)-entropy & Gate sensitivity \\
\midrule
AC-GATE        & $-0.0286\,(0.0050)$ & $0.108$ & $1.857$ & $0.269$ \\
No Recon Reg.  & $-0.0286\,(0.0050)$ & $0.108$ & $1.857$ & $0.269$ \\
No AC Encoder  & $-0.0284\,(0.0046)$ & $0.000$ & $1.845$ & $0.000$ \\
Uniform Lag    & $-0.0299\,(0.0035)$ & $0.000$ & $2.303$ & $0.394$ \\
Plain LSTM     & $-0.0288\,(0.0042)$ & $1.331$ & $2.059$ & --- \\
TFT            & $-0.0334\,(0.0055)$ & $1.730$ & $1.775$ & --- \\
GA-Net         & $-0.0331\,(0.0042)$ & $1.419$ & $1.993$ & --- \\
Grouped ARDL   & $\;\;0.6071\,(0.0000)$ & ---  & ---     & --- \\
\bottomrule
\end{tabular*}
\par\smallskip
{\footnotesize\raggedright\textit{Notes.} Cells: mean (sd) over 20 seeds. For AC-GATE variants, $k^\star_\sigma$ is computed from the learned lag-gate distribution. For Plain LSTM, TFT, and GA-Net, $k^\star_\sigma$ is a diagnostic lag-summary statistic computed for comparison and is not a structural gate-derived effective lag.\par}
\end{table}

\FloatBarrier
\clearpage
\subsection{L2 Stratified \(k^\star\) (All Stratifiers)}
\begin{table}[!tbhp]
\centering
\caption{Economics L2 stratified \(k^\star\) summary.}
\label{tab:app_econ_l2_full}
\scriptsize
\setlength{\tabcolsep}{3pt}
\noindent\begin{tabular*}{\columnwidth}{@{\extracolsep{\fill}} l l r r r r}
\toprule
Method & Stratifier & \(\overline{|\rho|}\) & median \(\rho\)
       & \(\Pr(p{<}5\%)\) & \(p_{\mathrm{Fisher}}\) \\
\midrule
AC-GATE & HC                         & $0.371$ & $-0.296$ & $0.80$ & $1.0{\times}10^{-46}$ \\
AC-GATE & log GDP per worker         & $0.278$ & $-0.146$ & $0.70$ & $3.5{\times}10^{-37}$ \\
AC-GATE & log capital per worker     & $0.257$ & $-0.135$ & $0.65$ & $1.4{\times}10^{-24}$ \\
No Recon Reg. & HC                   & $0.418$ & $-0.298$ & $0.80$ & $1.5{\times}10^{-50}$ \\
No Recon Reg. & log GDP per worker   & $0.287$ & $-0.126$ & $0.65$ & $2.5{\times}10^{-31}$ \\
No Recon Reg. & log capital per worker & $0.272$ & $-0.124$ & $0.60$ & $6.1{\times}10^{-30}$ \\
\bottomrule
\end{tabular*}
\par\smallskip
{\footnotesize\raggedright\textit{Notes.} No-AC-Encoder and Uniform-Lag are L1-degenerate (all 20 seeds invalid) and are omitted. Plain LSTM, TFT, and GA-Net may yield diagnostic lag summaries, but they do not produce structural gate-derived $k^\star$ values and are therefore excluded from L2 mechanism claims.\par}
\end{table}

\begin{table}[!tbhp]
\centering
\caption{Energy L2 stratified \(k^\star\) summary.}
\label{tab:app_energy_l2_full}
\scriptsize
\setlength{\tabcolsep}{3pt}
\noindent\begin{tabular*}{\columnwidth}{@{\extracolsep{\fill}} l l r r r r}
\toprule
Method & Stratifier & \(\overline{|\rho|}\) & median \(\rho\)
       & \(\Pr(p{<}5\%)\) & \(p_{\mathrm{Fisher}}\) \\
\midrule
AC-GATE & rule of law              & $0.735$ & $\;\;0.237$ & $0.95$ & $1.8{\times}10^{-79}$ \\
AC-GATE & gov.\ effectiveness      & $0.716$ & $\;\;0.129$ & $0.90$ & $9.0{\times}10^{-77}$ \\
AC-GATE & log GDP per capita       & $0.609$ & $\;\;0.185$ & $0.90$ & $1.5{\times}10^{-77}$ \\
No Recon Reg. & rule of law        & $0.735$ & $\;\;0.237$ & $0.95$ & $1.8{\times}10^{-79}$ \\
No Recon Reg. & gov.\ effectiveness& $0.716$ & $\;\;0.129$ & $0.90$ & $9.0{\times}10^{-77}$ \\
No Recon Reg. & log GDP per capita & $0.609$ & $\;\;0.185$ & $0.90$ & $1.5{\times}10^{-77}$ \\
\bottomrule
\end{tabular*}
\par\smallskip
{\footnotesize\raggedright\textit{Notes.} No-AC-Encoder and Uniform-Lag are L1-degenerate (all 20 seeds invalid) and are omitted. \(\overline{|\rho|}\) is sign-robust; the near-zero seed-mean of signed \(\rho\) reflects the seed-wise sign-flip invariance discussed in Section~\ref{sec:audit_protocol} (L2). Plain LSTM, TFT, and GA-Net may yield diagnostic lag summaries, but they do not produce structural gate-derived $k^\star$ values and are therefore excluded from L2 mechanism claims.\par}
\end{table}

\FloatBarrier
\section{Reproducibility}\label{app:repro}

\FloatBarrier
\subsection{Environment}
\begin{table}[!tbhp]
\centering
\caption{Experiment environment for the locked 20-seed suite.}
\label{tab:app_env}
\footnotesize
\setlength{\tabcolsep}{4pt}
\renewcommand{\arraystretch}{1.05}
\noindent\begin{tabular*}{\columnwidth}{@{\extracolsep{\fill}} l l}
\toprule
Item & Value \\
\midrule
OS                       & Windows 11 (10.0.26200) \\
Python                   & 3.13.9 \\
PyTorch                  & 2.9.1+cu130 \\
CUDA available           & yes \\
Device                   & \texttt{cuda} (NVIDIA GeForce RTX 5080) \\
\texttt{cudnn.deterministic} & false \\
\texttt{cudnn.benchmark}     & false \\
Deterministic algorithms & off \\
\bottomrule
\end{tabular*}
\end{table}

\FloatBarrier
\subsection{Data Preprocessing}
\begin{table}[!tbhp]
\centering
\caption{Data preprocessing summary.}
\label{tab:app_data_preprocessing}
\setlength{\tabcolsep}{4pt}
\renewcommand{\arraystretch}{1.05}
\noindent\begin{tabular*}{\columnwidth}{@{\extracolsep{\fill}} l l}
\toprule
Domain & Cleaning \\
\midrule
Synthetic & Balanced; no repair \\
Economics & \(ck, rgdpna>0\); miss.\ $\leq15\%$; lin.\ interp. \\
Energy & ISO3; pop,GDP\(>0\); miss.\ $\leq15\%$; lin.\ interp. \\
\bottomrule
\end{tabular*}
\end{table}

\subsection{GitHub link}
\noindent  \url{https://github.com/Oktinner0320/AC_GATE-CMDL.git}
\noindent Reproducible experiments were performed via notebooks, and results are located in the outputs/paper\_assets folder.

\FloatBarrier
\subsection{Experiment Parameters}
\begin{table}[!tbhp]
\centering
\caption{Locked experiment parameters.}
\label{tab:app_experiment_parameters}
\setlength{\tabcolsep}{4pt}
\renewcommand{\arraystretch}{1.05}
\noindent\begin{tabular*}{\columnwidth}{@{\extracolsep{\fill}} l l l}
\toprule
Scope & Block & Value \\
\midrule
Shared & Seeds & 20 (0--19) \\
Shared & Core & \(K=10\), \(d=64\), LSTM=2 \\
Shared & Optim. & drop 0.05; lr \(10^{-3}\); pat 20 \\
Shared & Lag gate & \(\lambda=0.1\); \(\tau=1\); softmax; b=1 \\
Shared & Runtime & clip 1; device auto \\
Syn. & Plan & lin+nonlin; ep 200; val 0.2 \\
Econ. & Contract & target ctfp; eff.-labor bundle \\
Econ. & Reg. & anchor recon; wt 2; detach=F \\
Econ. & Clip & split \\
Energy & Contract & ren. share $\rightarrow$ CO$_2$/energy; min bundle \\
Energy & Sched. & ep 120; clip=global \\
\bottomrule
\end{tabular*}
\end{table}

\FloatBarrier
\subsection{Synthetic DGP Details}
\label{app:synthetic_dgp}

\begin{algorithm}[!tbhp]
\caption{Synthetic data-generating process}
\label{alg:synthetic_dgp}
\Input{Entities \(N\), time steps \(T\), lag horizon \(K\), scenario \(r\in\{\mathrm{linear},\mathrm{nonlinear}\}\)}
\Output{Synthetic panel \(\{X_{i,t},p_i,s_i,Y_{i,t}\}\) and lag-center labels \(k_i^c\)}
\For{\(i=1,\ldots,N\)}{
    Draw latent moderator \(z_i\sim U(0,1)\)\;
    Set \(k_i^c=\mathrm{clip}(\mathrm{round}(3+7(1-z_i)),1,K)\) if \(r=\mathrm{linear}\)\;
    Set \(k_i^c=\mathrm{clip}(\mathrm{round}(10(1-z_i)^2),1,K)\) if \(r=\mathrm{nonlinear}\)\;
    Generate noisy proxy variables \(p_i\) from transformations of \(z_i\)\;
    Generate static features \(s_i\), with one dimension correlated with \(z_i\)\;
    Generate autoregressive inputs \(X_{i,t}\) with an entity shift proportional to \(z_i-0.5\)\;
    Build smooth DGP lag weights \(w^{\mathrm{DGP}}_{i,k}\) centered at \(k_i^c\)\;
    Generate \(Y_{i,t}\) from lag-weighted historical inputs, a small contemporaneous input term, entity-level terms, and Gaussian noise\;
}
\end{algorithm}
{\footnotesize\raggedright\textit{Notes.} Synthetic \(k^\star\) recovery is evaluated against the lag-center label \(k_i^c\). The synthetic setting is used as a controlled mechanism-recovery test rather than a general forecasting-superiority benchmark.\par}

\printbibliography
\vspace{12pt}
\end{document}